\documentclass[letterpaper, 10 pt, conference]{ieeeconf}  

\IEEEoverridecommandlockouts                              
\usepackage[latin1]{inputenc}
\usepackage[cmex10]{amsmath}
\usepackage{amsfonts}
\usepackage{amssymb}
\usepackage{amsbsy} 
\usepackage{hyperref}
\usepackage{comment}
\usepackage[scriptsize,tight]{subfigure}
\usepackage[]{units}
\usepackage{bigstrut}

\usepackage[pdftex]{graphicx}
\graphicspath{{./figures/}}
\DeclareGraphicsExtensions{.pdf}

\usepackage[disable]{todonotes}

\newenvironment{remark}[1][Remark]{\begin{trivlist} \item[\hskip \labelsep {\bfseries #1}]}{\end{trivlist}}

\usepackage{times} 

\title{\LARGE \bf Balancing experiments on a torque-controlled humanoid with hierarchical inverse dynamics}
\author{Alexander Herzog$^{1}$, Ludovic Righetti$^{1,2}$, Felix Grimminger$^{1}$, Peter Pastor$^{2}$ and Stefan Schaal$^{1,2}$
\thanks{\scriptsize This research was supported in part by National Science Foundation grants ECS-0326095, IIS-0535282, IIS-1017134, CNS-0619937, IIS-0917318, CBET-0922784, EECS-0926052, CNS-0960061, the DARPA program on Autonomous Robotic Manipulation, the Army Research Office, the Okawa Foundation, the ATR Computational Neuroscience Laboratories, and the Max-Planck-Society.}
\thanks{\scriptsize$^{1}$Autonomous Motion Department, Max Planck Institute for Intelligent Systems, %
T\"ubingen, Germany. {\tt\small first.lastname@tuebingen.mpg.de}}
\thanks{\scriptsize$^{2}$CLMC Lab, University of Southern California, Los Angeles, USA.}%
}
 
\renewcommand{\baselinestretch}{0.96}

\begin{document}

\maketitle
\thispagestyle{empty}
\pagestyle{empty}

\begin{abstract}
Recently several hierarchical inverse dynamics controllers based on cascades of quadratic
programs have been proposed
for application on torque controlled robots. 
They have important theoretical benefits but have never been implemented on
a torque controlled robot where model inaccuracies and real-time computation
requirements can be problematic.
In this contribution we present an experimental evaluation of these algorithms in the context of balance control for a humanoid robot.
The presented experiments demonstrate the applicability of the approach under real robot conditions (i.e. model uncertainty, estimation errors, etc).
We propose a simplification of the optimization problem that allows us to decrease computation
time enough to implement it in a fast torque control loop.
We implement a momentum-based balance controller which
shows robust performance in face of unknown disturbances, even when the robot is standing on only one foot.
In a second experiment, a tracking task is evaluated to demonstrate the performance
of the controller with more complicated hierarchies.
Our results show that hierarchical inverse dynamics controllers can be used
for feedback control of humanoid robots and that momentum-based balance control
can be efficiently implemented on a real robot.
\end{abstract}

\vspace{-0.1cm}
\section{INTRODUCTION}
We expect autonomous legged robots to perform complex tasks in persistent interaction with an uncertain and changing environment (e.g. in a disaster relief scenario). Therefore, we need to design algorithms that can generate precise but compliant motions while optimizing the interactions with the environment.
In this context, the choice of a control strategy for legged robots is of primary importance
as it can drastically improve performance in face of unexpected disturbances and therefore
open the way for agile robots, whether they are locomoting or performing manipulation tasks.

Robots with torque control capabilities~\cite{Hutter:2012uu,Boaventura:2012tc}, including humanoids~\cite{Cheng:2008tw,Ott:2011uj}, are becoming increasingly available and torque control algorithms are therefore necessary to fully
exploit their capabilities.
Indeed, such algorithms often offer high performance for motion control while
guaranteeing a certain level of compliance
\cite{Boaventura:2012tc,Kalakrishnan:2011dy,Saab:2011gg,Salini:2011bw}.
In addition, they also allow for the
direct control of contact interactions with the environment \cite{Hutter:2012uu,Righetti:2012uc,Righetti:2013tt},
which is required during operation in dynamic and uncertain environments.
Recent contributions have demonstrated the relevance of torque control approaches for humanoid robots,
for example for balancing capabilities \cite{Hyon:2007jya,Lee:2012hb,Ott:2011uj,Stephens:vu}.
All these approaches try to regulate the position of the center of mass (CoM) of the robot
to ensure that the robot does not fall while guaranteeing that contact forces are physically admissible.
We can essentially distinguish two approaches.

\textit{Passivity-based} approaches on humanoids were originally proposed in \cite{Hyon:2007jya}
and recently extended in \cite{Ott:2011uj}. They compute admissible contact forces and control commands under
quasi-static assumptions. The great advantage of such approaches is that they do not require a precise dynamic
model of the robot. Moreover, robustness is generically guaranteed due to the passivity property of the controllers.
However, the quasi-static assumption can be a limitation for dynamic motions.

On the other hand, controllers based on the \textit{full dynamic model} of the robot  
have also been successfully implemented on legged robots
\cite{Hutter:2012uu,Righetti:2013tt,Stephens:vu}. These methods essentially perform a form of inverse dynamics.
The advantage of such approaches is that they are in theory well suited for very dynamic motions. 
However, the need
for a precise dynamic model, sensor noise (particularly in the velocities) and limited torque bandwidth makes them more challenging to implement.
Moreover, it is generally required to simplify the
optimization process to meet time requirements of fast control loops (typically \unit[1]{kHz}
on modern torque controlled robots). Practical evaluations of both approaches are still rare due to the lack of torque controlled humanoid platforms 
and the complexity in conducting such robot experiments.

Recent approaches for controlling hierarchies of tasks using the full dynamics of the robot
are also very promising \cite{deLasa:2010hf, Saab:2011gg,Mansard:2012gy}.
Their problem formulations are more general than approaches based on pseudo-inverses as they
naturally allow the inclusion of arbitrary types of tasks including inequalities. The resulting optimization problems are phrased as cascades of quadratic programs (QPs).
Evaluation of their applicability was done in simulation and it has been argued that these algorithms can be implemented fast enough and that they can work on robots with model-uncertainty, sensor noise and limited torque bandwidth. But to the best of our knowledge, these controllers have never been used as feedback-controllers on real torque controlled humanoids. 
In \cite{Mansard:2012gy}, the trajectories computed in simulation are replayed on a real robot using joint space position control but the method is
not used for feedback control on the robot.
It is worth mentioning that \cite{Hutter:2012uu} recently successfully implemented a
controller using the full dynamics of the robot and task hierarchies on a torque controlled quadruped
robot. The approach is based on pseudo-inverses and not QPs which makes
it potentially inefficient to handle inequalities.

In this contribution we evaluate the balancing performance of a humanoid robot running hierarchical inverse dynamics controllers phrased as cascades of QPs~\cite{deLasa:2010hf}.
First we propose a simplification of the dynamic constraints that allow us to generally reduce the computational complexity of algorithms using inverse dynamics. It allows us to implement
our controller in a \unit[1]{kHz} real-time control loop.
Our main focus is then on presenting various balancing experiments in order to demonstrate the applicability of the approach under real robot conditions, i.e. model uncertainty, sensor noise, state estimation errors and a limited control bandwidth.
In one experiment, we implement a momentum-based balance controller as originally proposed
by \cite{Kajita:2003gj} and further developed in \cite{Lee:2012hb}
to take into account dynamic constraints. This experiment demonstrates the capabilities
of such momentum-based balance controllers on a torque controlled robot.
The second experiment is a tracking task, demonstrating that tracking accuracy in operational
space can still be achieved. 
It is, to the best of our knowledge, the first demonstration of the applicability of the methods
proposed in \cite{deLasa:2010hf} or \cite{Saab:2011gg} as feedback controllers on torque
controlled humanoids (i.e. without joint space PD control).

\section{HIERARCHICAL INVERSE DYNAMICS}\label{balance_controller}
For our experiments we write hierarchical inverse dynamics controllers and implement the solver proposed by~\cite{deLasa:2010hf} to perform real-time feedback control. In the following we give a short summary on how tasks can be formulated and revisit the original solver formulation. In Sect.~\ref{sec:decomp} we then propose a simplification to reduce the complexity of the original formulation. The simplification
is also applicable to any other inverse dynamics formulations.
\subsection{Modelling Assumptions and Problem Formulation} \label{sec:task_formulation}

Assuming rigid-body dynamics, we can write the equations of motion of a robot as
%
\begin{equation}\label{equations_of_motion}
\mathbf{M}(\mathbf{q}) \ddot{\mathbf{q}} + \mathbf{N}(\mathbf{q},\dot{\mathbf{q}}) = \mathbf{S}^T \boldsymbol{\tau} + \mathbf{J}_c^T \boldsymbol{\lambda}
\end{equation}
where $\mathbf{q} =[\mathbf{q}_j^T~\mathbf{x}^T]^T$ denotes the configuration of the robot. 
$\mathbf{q}_j \in \mathbb{R}^n$ is the vector of joint positions and $\mathbf{x} \in \mathrm{SE}(3)$
denotes the position and orientation of a frame fixed to the robot with respect to an inertial frame (the floating base).
$\mathbf{M}(\mathbf{q})$ is the inertia matrix, $\mathbf{N}(\mathbf{q},\dot{\mathbf{q}})$ is the vector
of all forces (Coriolis, centrifugal, gravity, friction, etc...), $\mathbf{S} = [\mathbf{I}_{n\times n} \mathbf{0}]$ represents the underactuation, $\boldsymbol{\tau}$ is the commanded joint torques, $\mathbf{J}_c$ is the Jacobian of the contact constraints and $\boldsymbol{\lambda}$
are the generalized contact forces.

Endeffectors in contact need to stay stationary. 
We express the constraint that the feet (or hands) in contact with the environment do not move ($\mathbf{x}_c = const$) by differentiating it twice and using the fact that $\dot{\mathbf{x}}_c = \mathbf{J}_c \dot{\mathbf{q}}$. We get the following equality constraint
\begin{equation}
\mathbf{J}_c \ddot{\mathbf{q}} + \dot{\mathbf{J}_c} \dot{\mathbf{q}} = \mathbf{0}.
\end{equation}

For the dynamics to be consistent, the centers of pressure (CoPs) at the endeffectors need to reside in the interior of each endeffector support polygon.
This can be expressed as a linear inequality by expressing the ground reaction force at the zero moment point.
For the feet not to slip we constraint the ground reaction forces (GRFs) to stay inside the friction cones. In our case, we
approximate the cones by pyramids to have linear inequality constraints.
Especially important for generating control commands that are valid on a robot, is to take into account actuation limits $\boldsymbol{\tau}_{min} \le \boldsymbol{\tau} \le \boldsymbol{\tau}_{max}$. The same is true for joint limits, which can be written as $\ddot{\mathbf{q}}_{min} \le \ddot{\mathbf{q}} \le \ddot{\mathbf{q}}_{max}$, where $\ddot{\mathbf{q}}_{min}, \ddot{\mathbf{q}}_{max}$ are computed from the distance of $\mathbf{q}$ to the joint limits at each instant of time.

Motion controllers can be phrased as $\ddot{\mathbf{x}}_{ref} = \mathbf{J}_x \ddot{\mathbf{q}} + \dot{\mathbf{J}}_x \dot{\mathbf{q}}$, where $\mathbf{J}_x$ is the task Jacobian 
and $\ddot{\mathbf{x}}_{ref}$ is a reference task acceleration (e.g. obtained from a PD-controller).
Desired contact forces can be directly expressed as equalities on $\boldsymbol{\lambda}$. 
In general, we assume that each control objective can be expressed as a linear combination of
$\ddot{\mathbf{q}}$, $\boldsymbol{\lambda}$ and $\boldsymbol{\tau}$.

At every control cycle, the equations of motion, Eq.~\eqref{equations_of_motion}, the constraints for physical consistency
(torque saturation, CoP constraints, etc...) and our control objectives will all be expressed
as affine equations of the variables $\ddot{\mathbf{q}}, \boldsymbol{\lambda}, \boldsymbol{\tau}$. Tasks of the same priority can then be stacked vertically into the form
%
\vspace{-0.3cm}
\begin{alignat}{2}
\label{affine_ctrl_objs_eq}
&\mathbf{A} \mathbf{y} + \mathbf{a}\leq 0, \\
\label{affine_ctrl_objs_ineq}
&\mathbf{B} \mathbf{y}+ \mathbf{b} = 0,
\end{alignat}
%
where $\mathbf{y} = [\ddot{\mathbf{q}}^T~\boldsymbol{\lambda}^T~\boldsymbol{\tau}^T]^T, \mathbf{A} \in \mathbb{R}^{m \times (2n+6+6c)}, \mathbf{a} \in \mathbb{R}^{m}, \mathbf{B} \in \mathbb{R}^{k \times (2n+6+6c)}, \mathbf{b} \in \mathbb{R}^{k}$ and $m, k \in \mathbb{N}$ the overall task dimensions and $n \in \mathbb{N}$ the number of robot DoFs. $c \in \mathbb{N}$ is the number of constrained endeffectors.

The goal of the controller is to satisfy these objectives as well as possible. Objectives will be stacked into different priorities, with the highest priority in the hierarchy given to physical consistency. In a lower priority we will express balancing and motion tracking tasks and we will put tasks for redundancy resolution in the lowest priorities.
\subsection{Hierarchical Inverse Dynamics Solver}
The control objectives and constraints in Eqs.~\eqref{affine_ctrl_objs_eq},\eqref{affine_ctrl_objs_ineq} might not have a common solution, but need to be traded off against each other. In case of a push, for instance, the objective to decelerate the CoG might conflict with a swing foot task. A tradeoff can be expressed in form of slacks on the expressions in Eqs.~\eqref{affine_ctrl_objs_eq},\eqref{affine_ctrl_objs_ineq}. The slacks then are minimized in a quadratic program
\vspace{-0.2cm}
\begin{alignat}{2}
\label{single_qp}
& \underset{\mathbf{y}, \mathbf{v}, \mathbf{w}}{\text{min.}}~
& 		 &\|\mathbf{v}\|^2 + \|\mathbf{w}\|^2 \\
\label{single_qp_ineq}
& \text{s.t.} & &~ \mathbf{V}(\mathbf{A} \mathbf{y} + \mathbf{a})\leq \mathbf{v}, \\
\label{single_qp_eq}
& 		 & & \mathbf{W}(\mathbf{B} \mathbf{y}+ \mathbf{b}) =  \mathbf{w},
\vspace{-0.1cm}
\end{alignat}
where matrices $\mathbf{V} \in \mathbb{R}^{m \times m}, \mathbf{W} \in \mathbb{R}^{k \times k}$ weigh the cost of constraints against each other and $\mathbf{v} \in \mathbb{R}^m, \mathbf{w} \in \mathbb{R}^k$ are slack variables. Note that $\mathbf{v}, \mathbf{w}$ are not predefined, but part of the optimization variables. In the remainder we write the weighted tasks as $\bar{\mathbf{A}}=\mathbf{V}\mathbf{A}, \bar{\mathbf{a}}=\mathbf{V}\mathbf{a}, \bar{\mathbf{B}}=\mathbf{W}\mathbf{B}, \bar{\mathbf{b}}=\mathbf{W}\mathbf{b}$.\\
Although, $\mathbf{W}, \mathbf{V}$ allow us to trade-off control objectives against each other, strict prioritization cannot be guaranteed with the formulation in Eq.~\eqref{single_qp}. 
For instance, we might want to trade off tracking performance of tasks against each other, but we do not want to sacrifice physical consistency of a solution at any cost.
In order to guarantee prioritization, we solve a sequence of QPs, where a QP with lower priority tasks is optimized over the set of optimal solutions of higher priority tasks as suggested by~\cite{deLasa:2010hf}. Given one solution $(\mathbf{y}_r^*, \mathbf{v}_r^*)$ for the QP of priority $r$, all remaining optimal solutions $\mathbf{y}$ in that QP are expressed by the equations 
\begin{eqnarray}
\vspace{-0.1cm}
\label{optimals_constr_1}
&\mathbf{y} = \mathbf{y}_r^* + \mathbf{Z}_r\mathbf{u}_{r+1}, \\
\label{optimals_constr_2}
&\bar{\mathbf{A}}_r \mathbf{y}+ \bar{\mathbf{a}}_r  \leq \mathbf{v}_r^*, \\ \nonumber
&\dots \\
&\bar{\mathbf{A}}_1 \mathbf{y}+ \bar{\mathbf{a}}_1 \leq  \mathbf{v}_1^*, \nonumber
\vspace{-0.0cm}
\end{eqnarray} 
where $\mathbf{Z}_r \in \mathbb{R}^{(2n+6+6c) \times z_r}$ represents a surjective mapping into the nullspace of all previous equalities $\bar{\mathbf{B}}_r, \dots,\bar{\mathbf{B}}_{1}$ and $\mathbf{u}_r \in \mathbb{R}^{z_r}$ is a variable that parameterizes that nullspace. In order to compute $\mathbf{Z}_r$ a Singular Value Decomposition (SVD) can be performed.
In our implementation it is performed in parallel with the QP at priority level $r-1$ and rarely finishes after the QP, such that it adds only a negligable overhead. 
 Now, we can express a QP of the next lower priority level $r+1$ and additionally impose the constraints in Eqs.~\eqref{optimals_constr_1},~\eqref{optimals_constr_2}, s.t. we can optimize over $\mathbf{y}$ without violating optimality of higher priority QPs:
\begin{alignat}{2}
\vspace{-0.1cm}
\label{hierarch_qp}
& \underset{\mathbf{u}_{r+1}, \mathbf{v}_{r+1}}{\text{min.}}~
& 		 &\|\mathbf{v}_{r+1}\| +  \|\bar{\mathbf{B}}_{r+1} (\mathbf{y}_r^* + \mathbf{Z}_r\mathbf{u}_{r+1}) + \bar{\mathbf{b}}_{r+1}\| \\
& ~~~\text{s.t.} 
&  &\bar{\mathbf{A}}_{r+1} ( \mathbf{y}_r^* + \mathbf{Z}_r\mathbf{u}_{r+1},) +\bar{\mathbf{a}}_{r+1} \leq \mathbf{v}_{r+1}, \nonumber \\
\label{hierarch_qp_previneq}
&&&~~~~~~\bar{\mathbf{A}}_r ( \mathbf{y}_r^* + \mathbf{Z}_r\mathbf{u}_{r+1},)+ \bar{\mathbf{a}}_r \leq \mathbf{v}_r^*,  \\
&&&~~~~~~\hspace{2cm} \dots \nonumber \\ 
&&&~~~~~~\bar{\mathbf{A}}_1 ( \mathbf{y}_r^* + \mathbf{Z}_r\mathbf{u}_{r+1},)+ \bar{\mathbf{a}}_1 \leq \mathbf{v}_1^*, \nonumber
\vspace{-0.1cm}
\end{alignat}
where we wrote the QP as in Eq.~\eqref{single_qp} and substituted $\mathbf{w}$ into the objective function. In order to ensure that we optimize over the optimal solutions of higher priority tasks, we added Eq.~\eqref{optimals_constr_2} as an additional constraint and substituted Eq~\eqref{optimals_constr_1} into Eqs.~\eqref{hierarch_qp}-\eqref{hierarch_qp_previneq}. This allows us to solve a stack of hierarchical tasks recursively as originally proposed by~\cite{deLasa:2010hf}. Note that this optimization algorithm is guaranteed to find the optimal solution in a least-squares sense while satisfying priorities.
%
%
\subsection{Decomposition of Equations of Motion}\label{sec:decomp}
Hierarchical inverse dynamics approaches usually have in common that consistency of the variables with physics, i.e. the equations of motion, need to be ensured. In~\cite{deLasa:2010hf} these constraints are expressed as equality constraints (with slacks) resulting in an optimization problem over all variables $\ddot{\mathbf{q}}, \boldsymbol{\tau}, \boldsymbol{\lambda}$. In~\cite{Mansard:2012gy} a mapping into the nullspace of Eq.~\eqref{equations_of_motion} is optained from a SVD on Eq.~\eqref{equations_of_motion}. In both cases complexity can be reduced as we will show in the following. We decompose the equations of motion as 
\begin{eqnarray}\label{decomp_eq_motion_upper}
\mathbf{M_u(\mathbf{q})} \ddot{\mathbf{q}} +
\mathbf{N_u}(\mathbf{q},\dot{\mathbf{q}}) 
&=& 
\boldsymbol{\tau} +
\mathbf{J}_{c,u}^T 
\boldsymbol{\lambda}, \\
\label{decomp_eq_motion_lower}
\mathbf{M_l(\mathbf{q})} \ddot{\mathbf{q}}  +
\mathbf{N_l}(\mathbf{q},\dot{\mathbf{q}})
&=&
\mathbf{J}_{c,l}^T
\boldsymbol{\lambda}
\end{eqnarray}
where Eq.~\eqref{decomp_eq_motion_upper} is just the first $n$ equations of Eq. \eqref{equations_of_motion} and eq.~\eqref{decomp_eq_motion_lower} is the last $6$ equations
related to the floating base. 
The latter can then be interpreted as the Newton-Euler equations of the whole system \cite{Wieber:2006up}. They express the change of momentum of the robot as a function
of external forces.
A remarkable feature of the decomposition in Eqs.~\eqref{decomp_eq_motion_upper},~\eqref{decomp_eq_motion_lower} is that the torques $\boldsymbol{\tau}$ only occur in Eq.~\eqref{decomp_eq_motion_upper} and are exactly determined by $\ddot{\mathbf{q}}, \boldsymbol{\lambda}$ in the form 
\begin{equation} \label{torque_substitution}
\boldsymbol{\tau} = \mathbf{M_u(\mathbf{q})}\ddot{\mathbf{q}} + \mathbf{N_u}(\mathbf{q},\dot{\mathbf{q}}) - \mathbf{J}_{c,u}^T \boldsymbol{\lambda}
\end{equation}
Since $\boldsymbol{\tau}$ is linearly dependent on $\ddot{\mathbf{q}}, \boldsymbol{\lambda}$,
for any combination of accelerations and contact forces there always exist a solution for $\boldsymbol{\tau}$. It is given by Eq.~\eqref{decomp_eq_motion_upper}.
Therefore, it is only necessary to use Eq.~\eqref{decomp_eq_motion_lower} 
as a constraint for the equations of motion during the optimization 
(i.e. the evolution of momentum is the only constraint).

Because of the linear dependence, all occurrences of $\boldsymbol{\tau}$ in the problem formulation (i.e. in Eqs.~\eqref{affine_ctrl_objs_eq}-\eqref{affine_ctrl_objs_ineq}) can be substituted by Eq.~\eqref{torque_substitution}.
This reduces the number of variables in the optimization from $(2n+6+6c)$ to $(n+6+6c)$. This decomposition thus saves as many variables as there are DoFs on the robot. 
This simplification is crucial to reduce the time taken by the optimizer and allowed
us to implement the controller in a \unit[1]{kHz} feedback control loop. 
\begin{remark}
The simplification that we propose\footnote{We originally proposed the simplification in this technical note~\cite{herzog:2013a}.} can appear trivial at first sight. However, it is worth mentioning that
such a decomposition is always ignored in related work despite the need for computationally fast 
algorithms~\cite{deLasa:2010hf},~\cite{Mansard:2012gy},~\cite{Stephens:vu}.
\end{remark}

\section{EXPERIMENTAL SETUP}\label{experimental_setup}
We now detail the experimental setup, the low-level feedback torque control and the limitations
of the hardware.  
%
%
\subsection{Sarcos Humanoid Robot}
\begin{figure}
\centering
\includegraphics[width=0.2\textwidth]{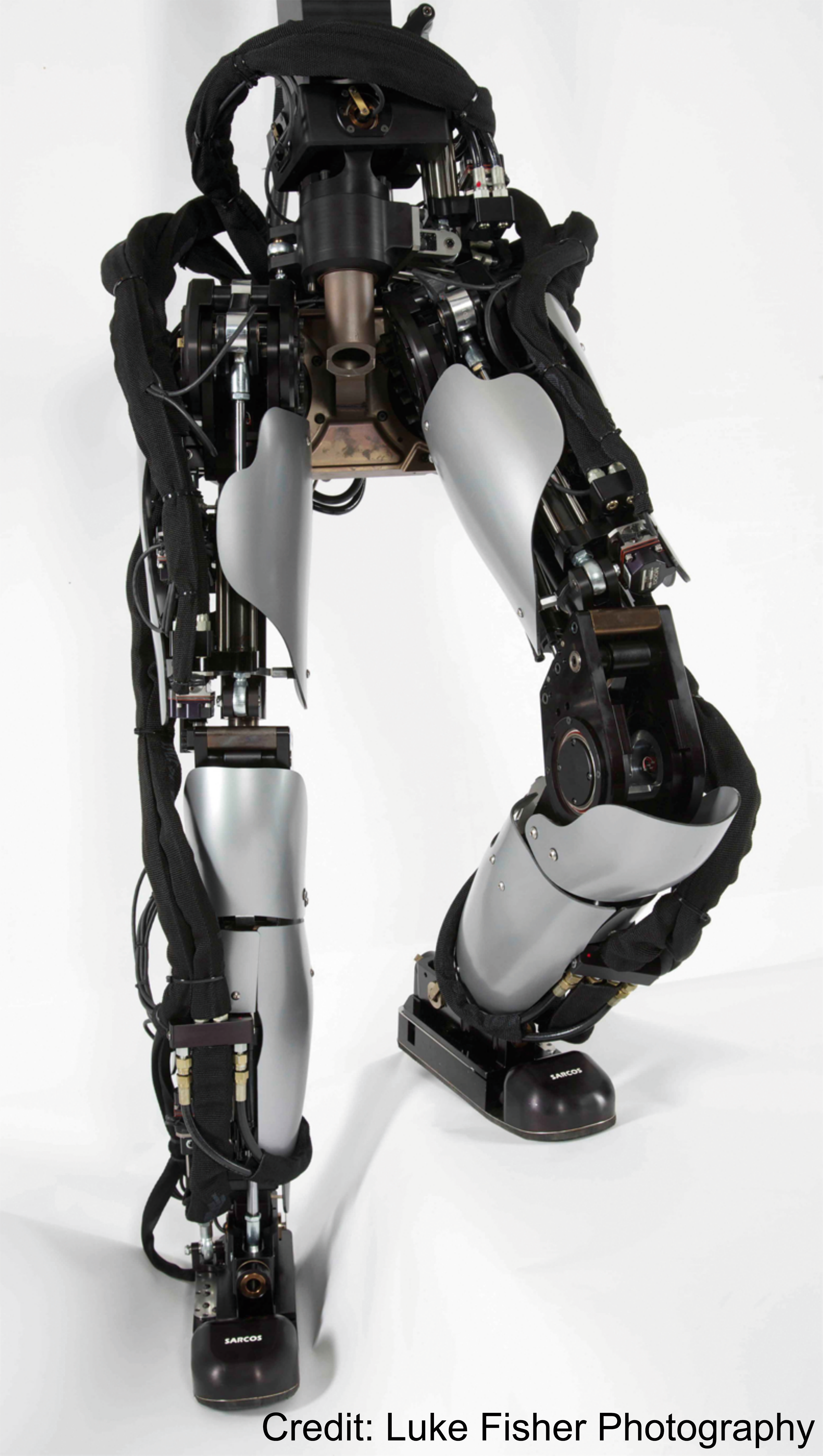}
\vspace{-0.2cm}
\caption{The lower part of the Sarcos Humanoid.}\label{fig:biped}
\vspace{-0.5cm}
\end{figure}
The experiments were done on the lower part of the Sarcos Humanoid Robot \cite{Cheng:2008tw} (Fig. \ref{fig:biped}).
It consists of two legs and a torso. The legs have 7 DOFs each and the torso has 3 DOFs.
Given that the torso supports a negligible mass because it is not connected to the upper body
of the robot and its motion does not change much the dynamics, we froze these DOFs during the experiments.
The legs of the robot are 0.82m high. Each foot is 0.09m wide and 0.25m long, which is rather small for a biped.
Note also that the front of the foot is made of a passive joint that is rather stiff, located 10cm before the tip of the foot.
The total robot mass is 51kg.
The robot is actuated with hydraulics and each joint consists of a Moog Series 30 flow control servo valve
with a piston on which a load cell is attached to measure the force at the piston.
A position sensor is also located at the joint.
Each foot has a 6-axis force sensor and we mounted an IMU on the pelvis of the robot
from which we measure angular velocities, linear accelerations and the orientation of the robot
in an inertial frame. An offboard computer sends control commands to the robot at \unit[1]{kHz}.
The control commands consist of the desired current applied to each valve.

\subsection{Low-level torque control}
For each actuator, we implemented a torque feedback controller that ensures that each joint
produces the desired force generated by the balance controller. The controller essentially
computes desired valve commands given a desired torque, the valve commands representing
a desired flow.
The controller we implemented is very much inspired from the work in \cite{Boaventura:2012tc, Boaventura:2012va}, with the difference
that we implemented a simpler version where piston velocity feedback has a constant gain.
The control law is
\begin{equation}
v = PID(F_{des},F) + K \dot{x}_{piston} + d
\end{equation}
where $v$ is the valve command, $PID$ is a PID control according to desired force command
and force measured from the load cells, $K$ is a positive gain, $\dot{x}_{piston}$ is the
piston velocity (computed from the joint velocity) and $d$ is a constant bias.

\begin{figure}
\vspace{-0.2cm}
\centering
\includegraphics[width=0.5\textwidth]{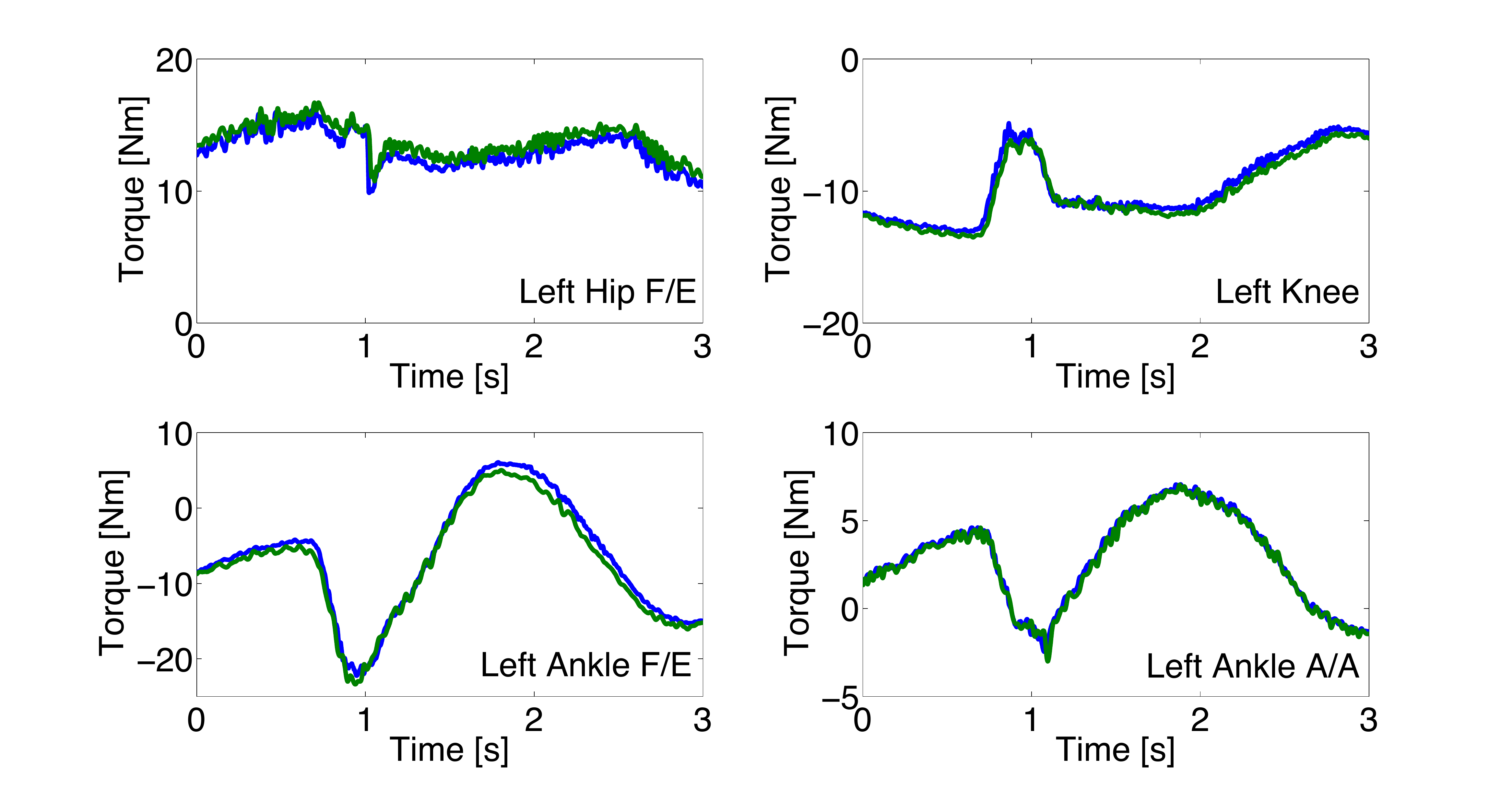}
\vspace{-0.7cm}
\caption{Example of torque tracking performance during a balancing experiment. The left hip flexion/extension, left knee and left ankle flexion/extension and adduction/abduction joints are shown. Both desired (blue) and actual (green) torques are shown.}\label{fig:torque_tracking}
\vspace{-0.5cm}
\end{figure}

This controller design allowed us to achieve good torque tracking performance.
It is important to note that it was necessary to achieve good performance in 
the hierarchical inverse dynamics controller.
Fig.~\ref{fig:torque_tracking} illustrates the torque tracking performance during the balancing experiment.

\subsection{Limitations}
During the experiments, we suffered from two important limitations.
First, our dynamic model is based on the CAD model of the robot. It means that it is not very accurate
as it does not take into account the contribution of the hydraulic hoses, the electronics nor any
type of friction in the model.
Second, we relied on very simple state estimation from the floating base.
We just integrated the acceleration
readings to get the base velocity and position, using a decay term to damp velocities.

However, these limitations are not too problematic as we show in the experiments.
Furthermore, they can
be overcome by using system identification \cite{Mistry:2009dh} and filtering \cite{Bloesch:2012wu} methods that have proven to work well on legged robots.

\section{EXPERIMENTS}
%
%
We formulated balancing and motion tracking tasks in the framework discussed in Sect.~\ref{balance_controller} and evaluated them on the Sarcos Humanoid described in Sect.~\ref{experimental_setup}. The performance of the controller was evaluated in three different scenarios: balancing experiments in single and double support and a tracking task. A summary of the experiments is shown in the attached movie
\footnote{An extended version is available on http://youtu.be/RlVOaW4vPU8}. 

For all the experiments, we run the hierarchical inverse dynamics controller as a feedback controller
that computes directly the desired torque commands. We do not use any joint PD controller for stabilization (i.e. feedback control is only done in task space).

\subsection{Processing Time}
\begin{figure}
\vspace{-0.35cm}
\centering 
\includegraphics[width=0.9\linewidth]{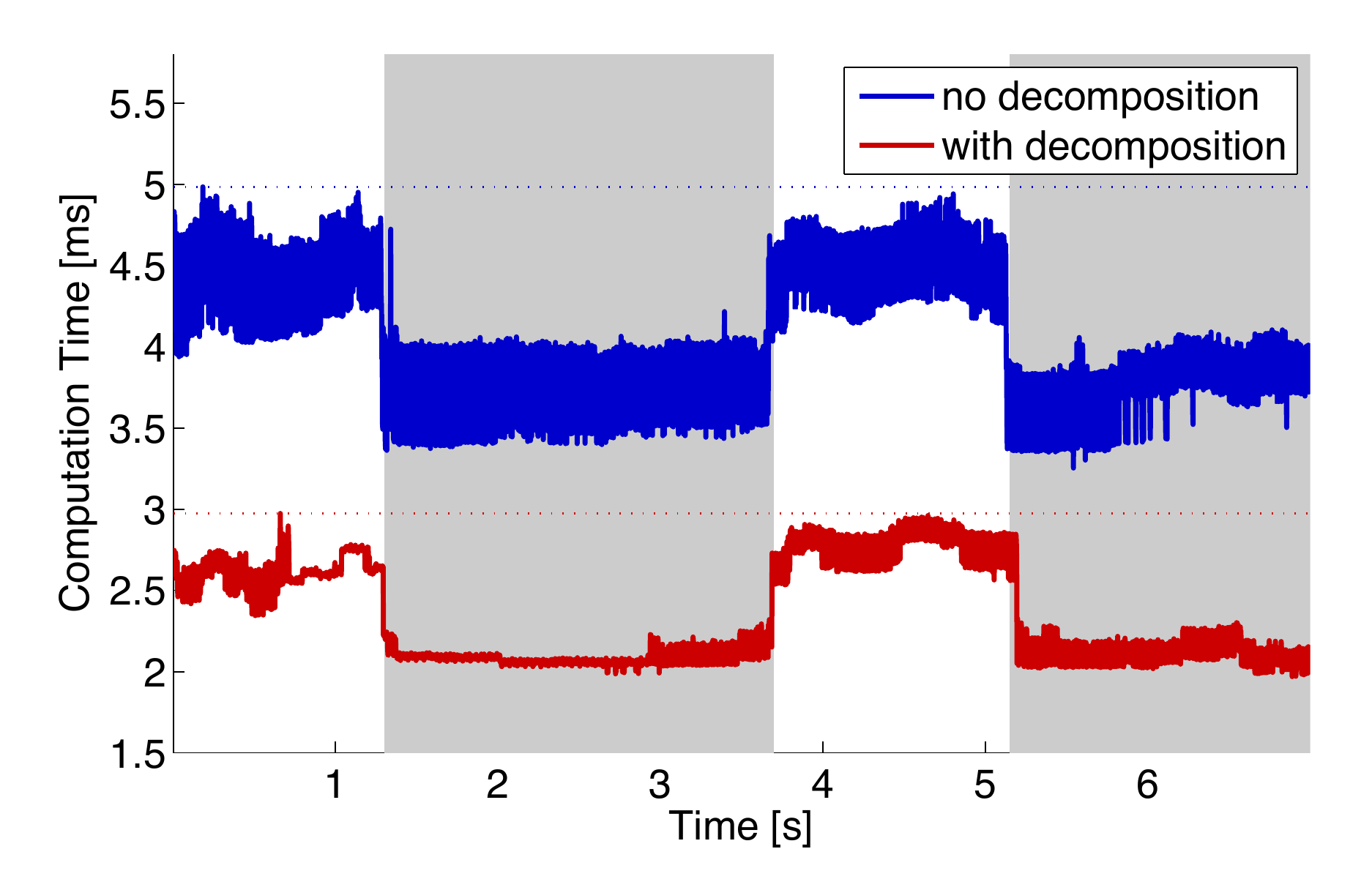}
\vspace{-0.5cm}
\caption{Processing time of a stepping task (see Table~\ref{tab:stepping_task}) using the decomposition proposed in Sect.~\ref{sec:decomp}~(red) and the same task performed without the decomposition (blue). The dotted line represents the maximum computation per control cycle respectively. Intervals shaded in gray show the robot in single support phase. In the remaining time the robot is in double support. With the proposed decomposition we decreased the computation time by approximately 40\%.}
\label{fig:comp_duration} 
\vspace{-0.5cm}
\end{figure}

The computation time of our solver mainly depends on a) the DoFs of the robot b) the number of contact constraints and c) the composed tasks. All experiments were performed on an Intel Core i7-2600 CPU with 3.40GHz.  In our experiments we will use the 14 DoFs lower part of a humanoid to perform several tasks in a \unit[1]{kHz} control loop. In the following, however, we construct a more complicated stepping task for the full 25 DoFs robot in order to evaluate the speedup from the simplification in Sect.~\ref{sec:decomp}.
\begin{table}[thp]
\center
\begin{tabular}{c c l}
  \hline
  \textbf{Priority} & \textbf{Nr. of eq(uality) and} & \textbf{Constraint/Task} \\
	&  \textbf{ineq(uality) constraints} &\\
  \hline
  1 &{\color{red} $25$ eq}& {\color{red}Eq.~\eqref{decomp_eq_motion_upper} (not required for} \\&&{\color{red}simplified problem)}\\
     &$6$ eq& Newton Euler Eq.~\eqref{decomp_eq_motion_lower}\\
     &$2\times 25$ ineq& torque limits\\
 2 &$c \times 6$ eq& kinematic contact constraint\\
     &$c\times 4$ ineq& CoPs reside in sup. polygons\\
     &$c\times 4$ ineq& GRFs reside in friction cones\\
     &$2 \times 25$ ineq& joint acceleration limits\\
 3 &$3$ eq& PD control on CoM\\
     &$(2-c)\times 6$& PD control on swing foot\\
  4 &$25+6$ eq& PD control on posture\\
  5 &$c \times 6$ eq& regularizer on GRFs\\
  \hline \hline
&\textbf{DoFs:} 25 &\textbf{max. time:}~{\color{red}  \unit[5]{ms}} /  \unit[3]{ms} \bigstrut[t]\\
\end{tabular}
\vspace{-0.0cm}
\caption{\small Full Humanoid Stepping Task for Speed Comparison}
\label{tab:stepping_task}
\vspace{-0.5cm}
\end{table}
As summarized in Table~\ref{tab:stepping_task} the highest two priorities satisfy hardware limitations and dynamic constraints, the third priority tracks the task relevant center of mass and swing foot motion and the remaining priorities resolve redundancies on motion and forces. The problem size changes depending on the number of contacts $c$ ($c=2$ in double support and $c=1$ in single support). The proposed decomposition removed 25 equality constraints and 25 optimization variables. We measured the computation time of both versions of the hierarchical solver, one with the full EoM and one with the proposed reduction as plotted in Fig.~\ref{fig:comp_duration}. Looking at the worst case (as this is significant for execution in a time critical control loop) we can drop computation time by 40\%. In our experiments with a 14 DoF robot, this speedup allows us to run a \unit[1]{kHz} control-loop as we will demonstrate in the following sections. It would not have been possible
without the simplification.
 In our speed comparison in Fig.~\ref{fig:comp_duration} one can see that computation time varies with the number of constrained endeffectors, which can be problematic if the number of contacts increases too much (e.g. when using both hands and feet).
  
\subsection{Balance Control Experiments}\label{sec:balance_ctrl}
Our first experiment is the implementation of a momentum balance controller
as originally proposed in ~\cite{Kajita:2003gj} and then extend in~\cite{Lee:2012hb}.
The idea is to regulate both the linear and angular momentum of the robot to keep balance.
In this implementation, the physical constraints are put in the highest priority.
In the second priority, we put the balancing task as well as the force regularization and
postural control. The task composition is summarized in Table~\ref{tab:momentum_task}.

In \cite{Orin:2008ge} a linear mapping 
\begin{equation}
\vspace{-0.1cm}
\label{momentum_mat_eq}
\mathbf{H_G}(\mathbf{q})\dot{\mathbf{q}} = \mathbf{h} 
\end{equation}
was derived that maps joint velocities to $\mathbf{h} = [\mathbf{h}_{lin}^T ~\mathbf{h}_{ang}^T]^T $, the system linear and angular momentum expressed at the CoM. The matrix $\mathbf{H_G}$ is called the centroidal momentum matrix. It was applied in~\cite{Lee:2012hb} to regulate the momentum by computing a desired momentum rate of change via PD control
\begin{eqnarray}\label{h_ref}
\dot{\mathbf{h}}_{ref} &=& 
\mathbf{P}
\begin{bmatrix}
m (\mathbf{x}_{cog,des} - \mathbf{x}_{cog}) \\ \mathbf{0}
\end{bmatrix} + \nonumber \\
&& \mathbf{D} (\mathbf{h}_{des} - \mathbf{h})
+ \dot{\mathbf{h}}_{des}
\end{eqnarray}
where $\mathbf{P}$ and $\mathbf{D}$ are positive-definite gain matrices. 
Typically, this can be used to regulate the position of the center of mass while
damping its linear velocity and the angular momentum. The derivative of Eq.~\eqref{momentum_mat_eq} allows us to express a controller on the system momentum
\begin{eqnarray}
 \dot{\mathbf{h}} &=& \mathbf{H_G}\ddot{\mathbf{q}} + \dot{\mathbf{H}}_\mathbf{G}\dot{\mathbf{q}} \label{mom_rate_ddq} \\
&=& \begin{bmatrix}
\mathbf{I}_{3\times 3} & \mathbf{0}_{3 \times 3} &\ldots	
\\
\label{mom_rate_frcs}
[\mathbf{x}_{cog} - \mathbf{x}_i]_{\times} &\mathbf{I}_{3\times 3}& \ldots	
\end{bmatrix}
\boldsymbol{\lambda} + 
\begin{bmatrix}
m\mathbf{g}
\\
\mathbf{0}
\end{bmatrix},
\end{eqnarray}
where $m\mathbf{g}$ is the gravitational force applied at the CoM and $[\centerdot]_{\times}$ maps a vector to a skew symmetric matrix, s.t. $[\mathbf{x}]_{\times}\boldsymbol{\lambda} = 
\mathbf{x} \times \boldsymbol{\lambda}$. Eq.~\eqref{mom_rate_frcs} is the general formula for the \textit{change of momentum} of a rigid multi-body system. 
We see that we can express the rate of momentum change either as a 
function of  $\mathbf{\ddot{q}}$ as in Eq.~\eqref{mom_rate_ddq} or as a function of
$\boldsymbol{\lambda}$ as in Eq.~\eqref{mom_rate_frcs}. We can express this task
either as a force task or as a kinematic task (the matrix in front of $\mathbf{\ddot{q}}$
or $\boldsymbol{\lambda}$ being the Jacobian of the task).
In this experiment we use the kinematic representation and in the tracking experiment
we use the force representation.
We note that in general, expressing the momentum control with Eq.~\eqref{mom_rate_frcs} might be
better because we do not have to compute $\dot{\mathbf{H}}_G$, which usually is acquired through numerical derivation and might suffer from magnified noise.

\begin{table}[]
\center
\begin{tabular}{c c l}
  \hline
  \textbf{Priority} & \textbf{Nr. of eq(uality) and} & \textbf{Constraint/Task} \\
	&  \textbf{ineq(uality) constraints} &\\
  \hline
  1 &$6$ eq& Newton Euler Eq.~\eqref{decomp_eq_motion_lower}\\
     &$2\times 14$ ineq& torque limits\\
     &$2 \times 6$ eq& kinematic contact constraint\\
     &$2\times 4$ ineq& CoPs reside in sup. polygons\\
     &$2\times 4$ ineq& GRFs reside in friction cones\\
     &$2 \times 14$ ineq& joint acceleration limits\\
 2 &$6$ eq& PD control on system \\&&momentum, Eq.~\eqref{h_ref}\\
    &$14+6$ eq& PD control on posture\\
    &$2 \times 6$ eq& regularizer on GRFs\\
  \hline \hline
&\textbf{DoFs:} 14 &\textbf{max. time:}~ \unit[0.4]{ms} \bigstrut[t]\\
\end{tabular}
\vspace{-0.0cm}
\caption{\small Hierarchy of Double Support Balancing Task}
\vspace{-0.9cm}
\label{tab:momentum_task}
\end{table}

In our first experiment we pushed the robot impulsively with a stick at various contact points with different force directions. 
To ensure that the robot is really balancing, the same experiment was conducted when running 
a simple inverse dynamics algorithm with
contact forces optimization~\cite{Righetti:2012uc}. As expected, the balance controller
showed a better balancing performance and did not fall over as it was the case for~\cite{Righetti:2012uc} as can
be seen in the video. When pushing the robot with a constant force at various parts, it stayed in balance and adapted its posture in a compliant manner. 
We also tested the controller when the feet were not co-planar, but one foot was put on top of a block as can
be seen in the movie.

\begin{figure}
\vspace{-0.2cm}
\centering
\begin{minipage}{0.45\textwidth}
\includegraphics[width=\textwidth]{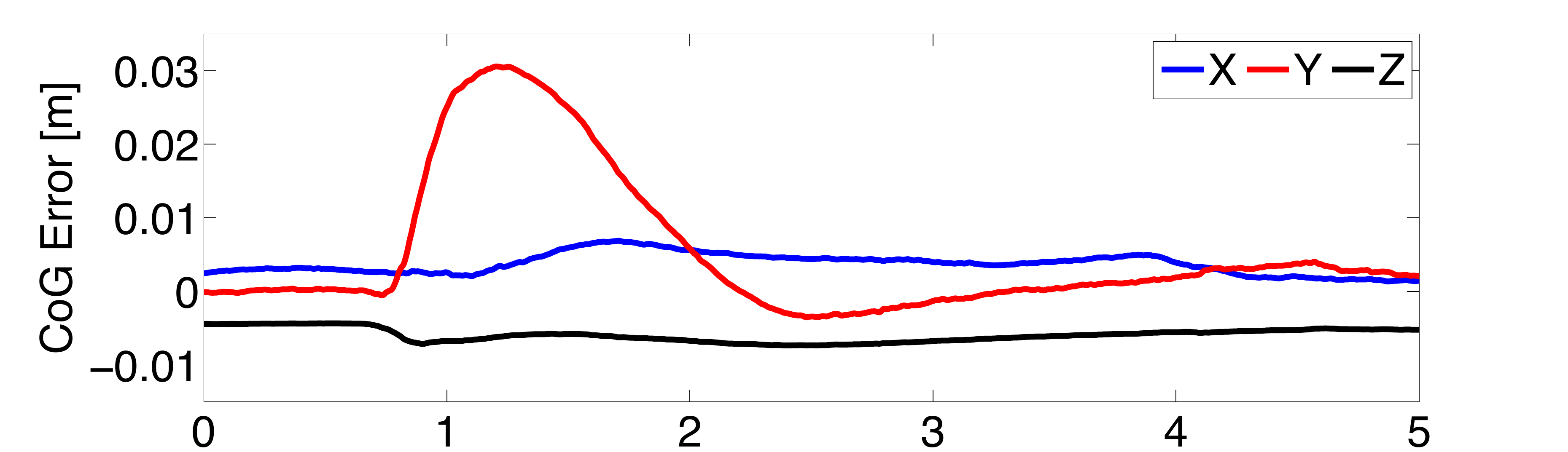}\\
\includegraphics[width=\textwidth]{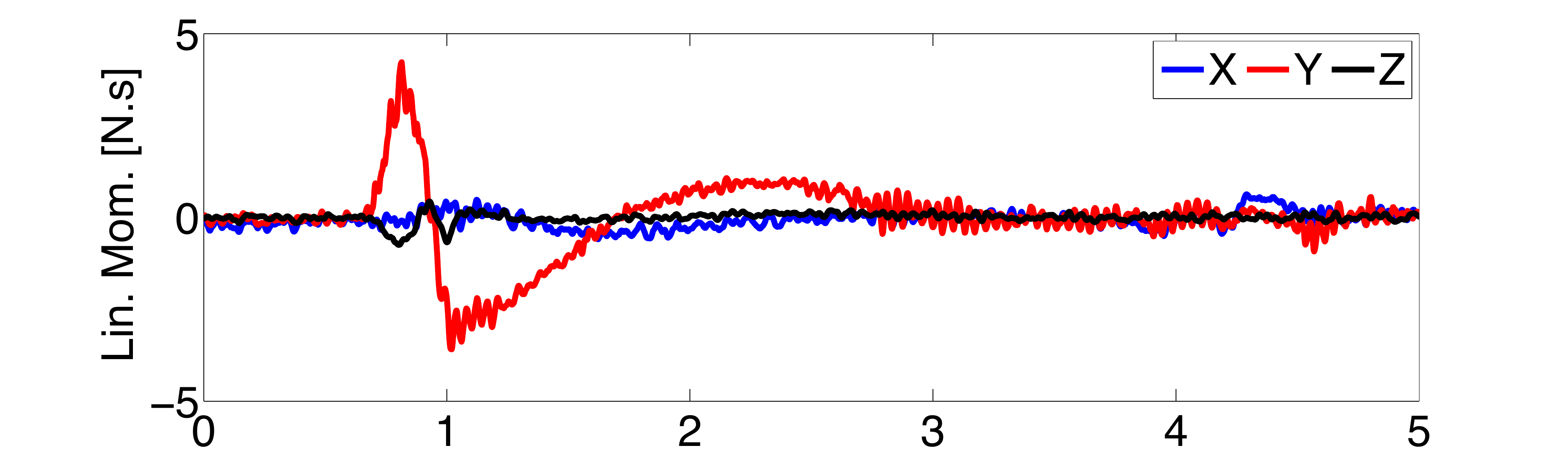}\\
\includegraphics[width=\textwidth]{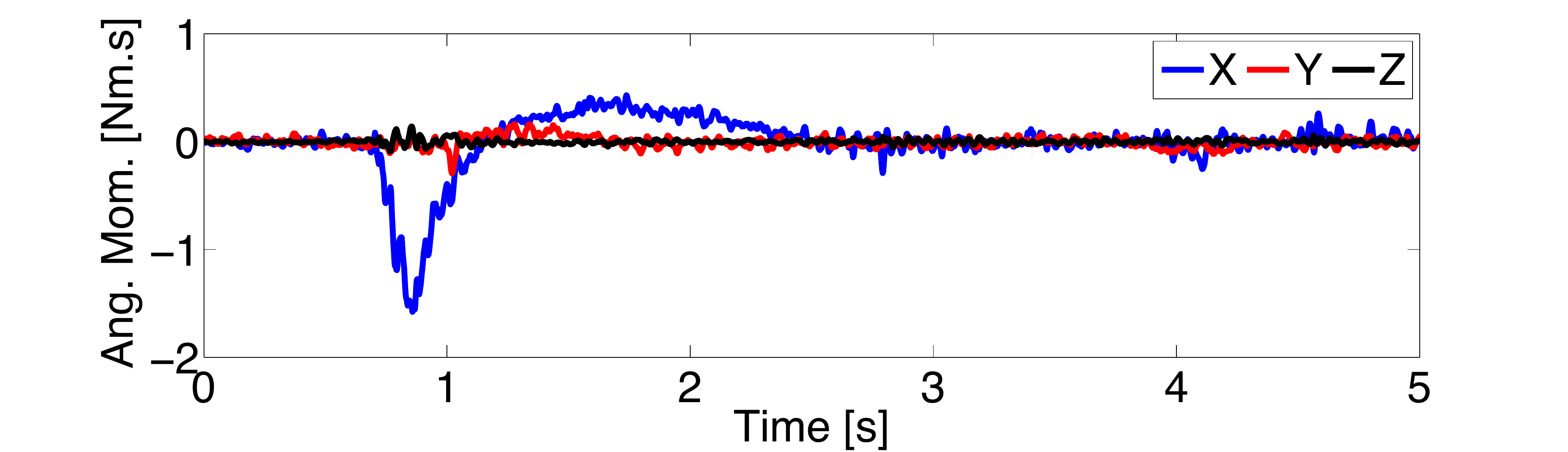}
\end{minipage}
\begin{minipage}{0.35\textwidth}
\includegraphics[width=\textwidth]{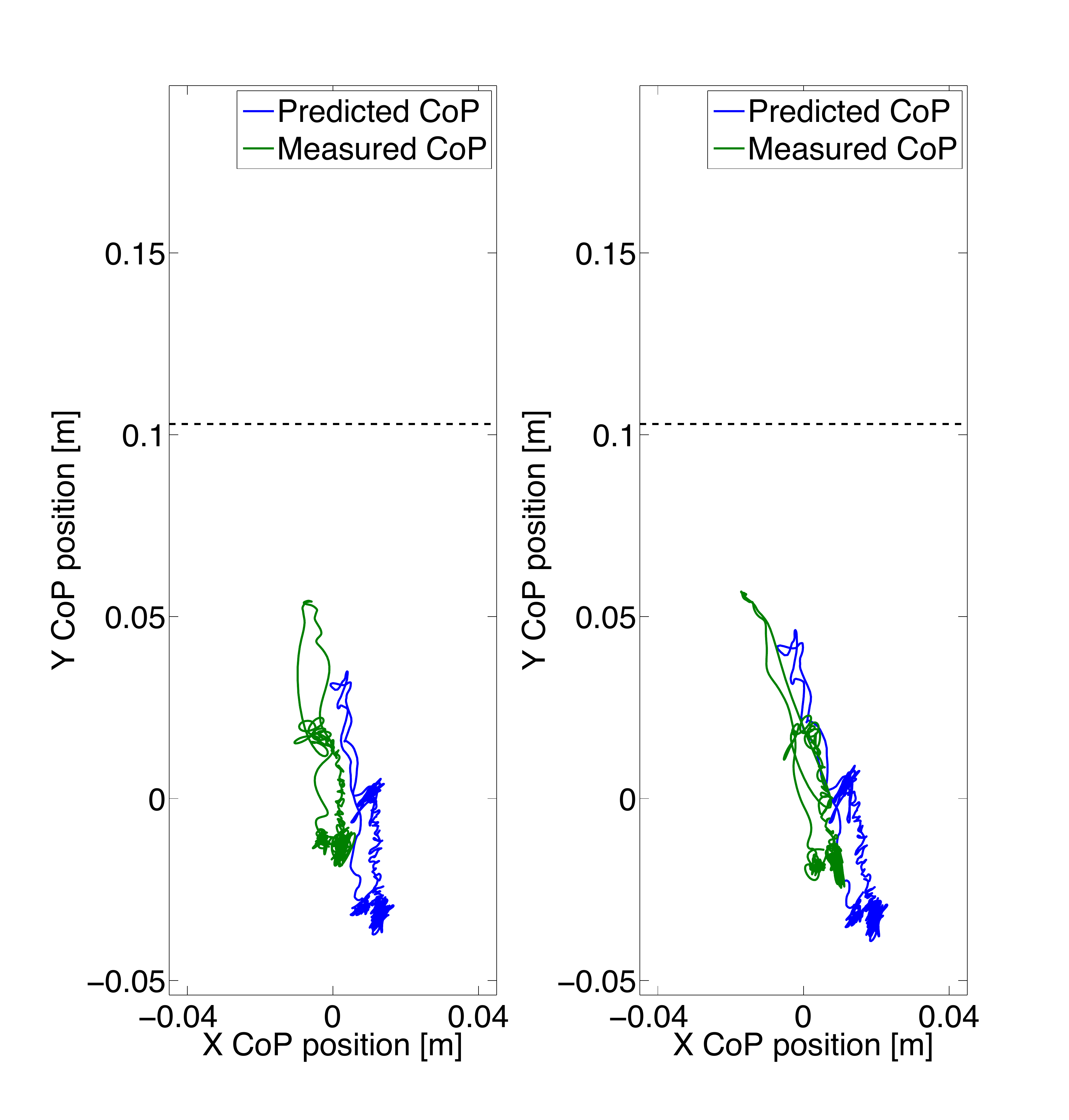}
\end{minipage}
\vspace{-0.1cm}
\caption{This figure shows the recovery of desired CoG position, linear and angular momentum after a push from the front.  X points in the robot right direction, Y in the forward direction and Z in the vertical direction. The change in CoP for each leg is also plotted in the lower graph. The plot has the same ratio as the feet and the horizontal dashed line shows the position of the passive joint at the front of the foot.}\label{fig:push_momentum}
\vspace{-0.5cm}
\end{figure}

In the planar posture, even pushes with a rather high magnitude were absorbed and the robot kept standing. The change in momentum was damped out quickly and the CoM was tracked after an initial disturbance as can be seen in Fig.~\ref{fig:push_momentum}. The CoPs remained inside of the support polygons and were tracked well.
We notice from Fig.~\ref{fig:push_momentum} that the CoPs, as they were predicted from the dynamics model, are approximately correct (within 2cm error).
However, one can expect that a higher precision might be needed to achieve dynamic motions which could be
achieved with an inertial parameters estimation procedure \cite{Mistry:2009dh}.

For our next experiment we put the biped on a rolling platform and rotated and moved it with a rather fast change of directions. Although, the CoM was moving away from its desired position initially, the momentum change was damped out and the robot kept standing and recovered CoM tracking. The stationary feet indicated that forces were applied that were consistent with our CoP boundaries.

\begin{figure}
\centering
\includegraphics[width=0.5\textwidth]{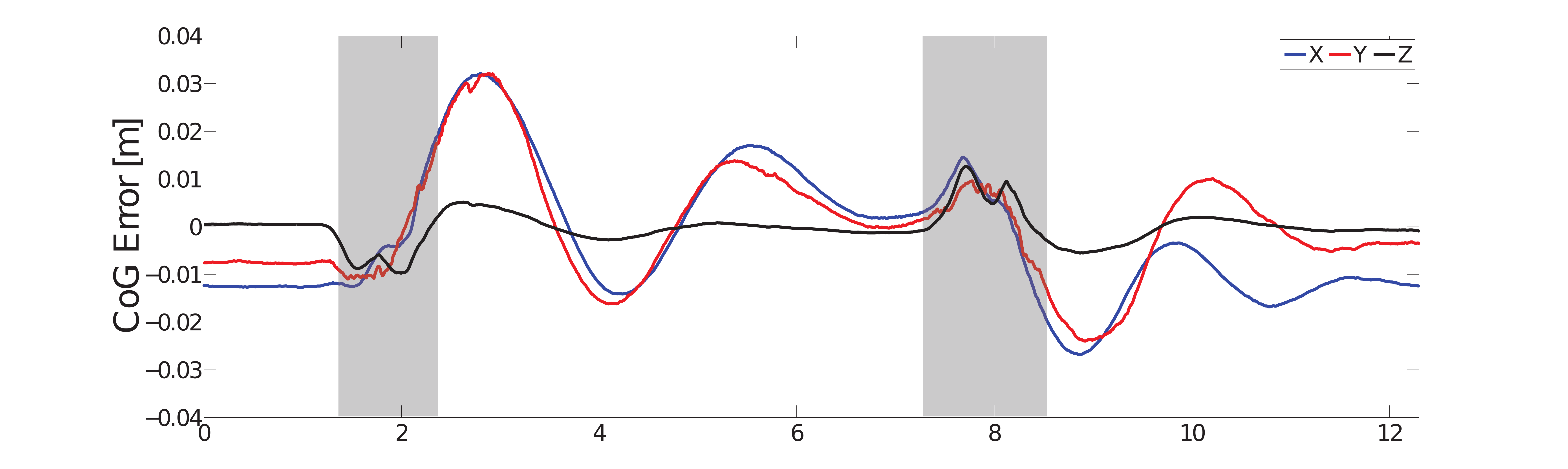}\\
\includegraphics[width=0.5\textwidth]{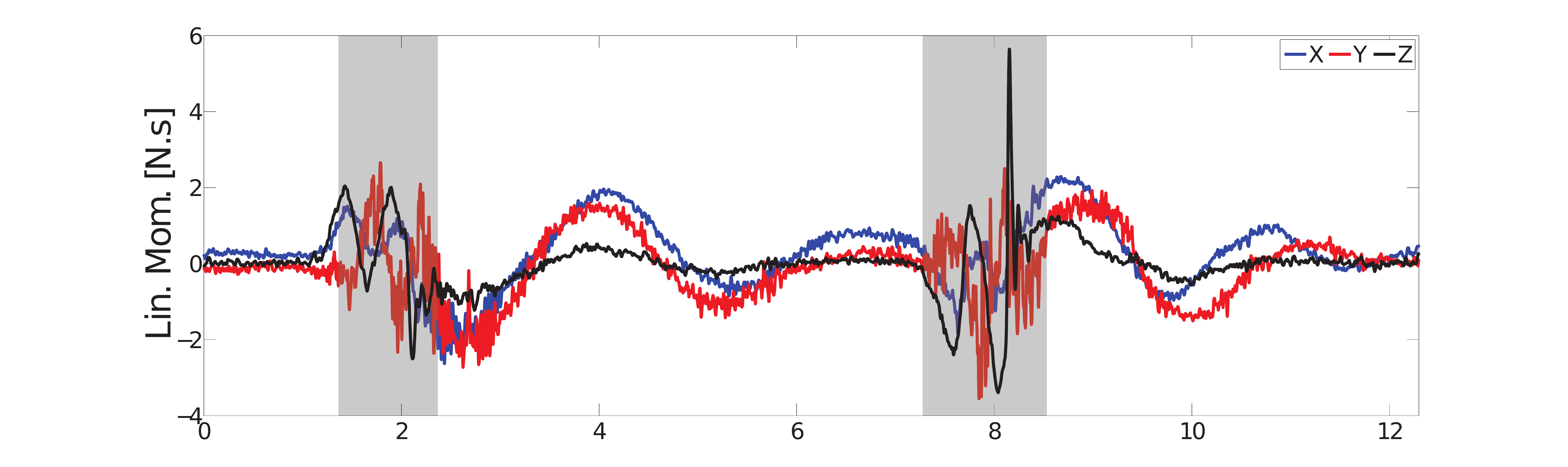}\\
\includegraphics[width=0.5\textwidth]{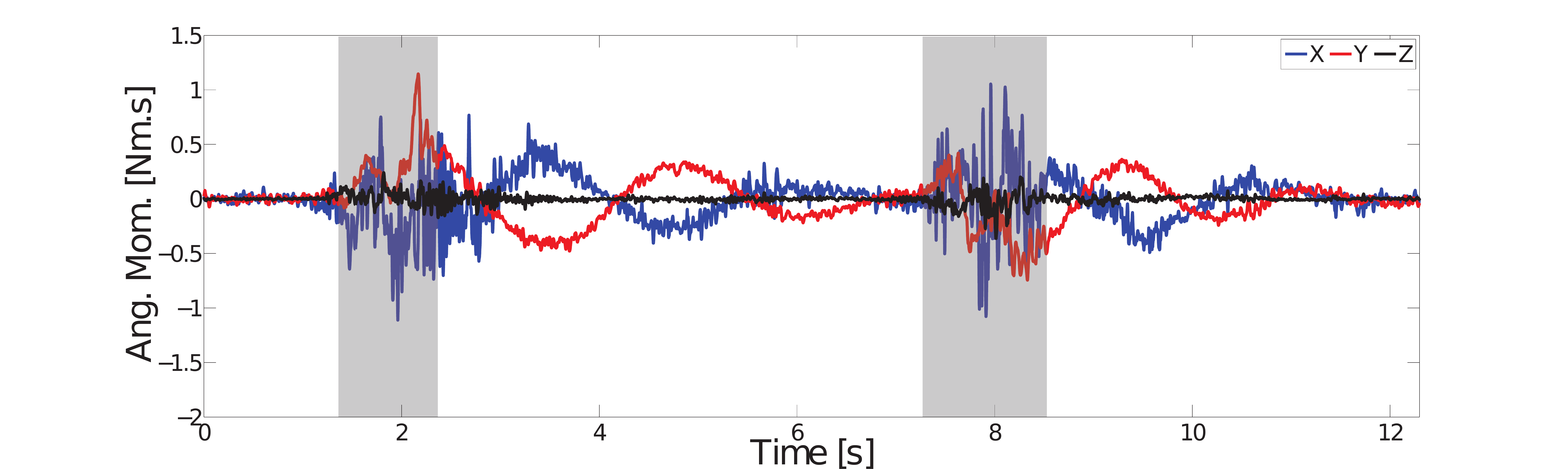}\\
\includegraphics[width=.35\textwidth]{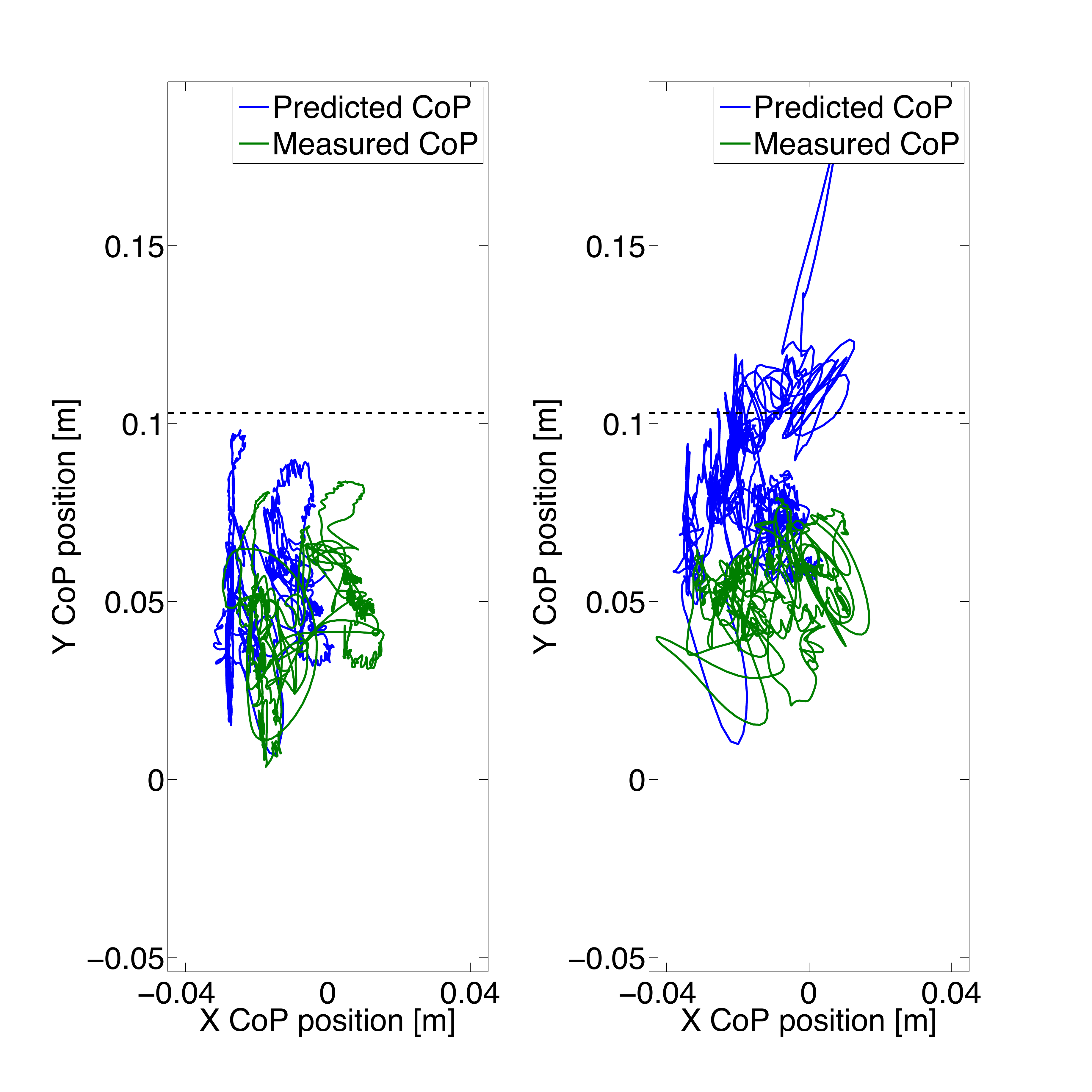}
\vspace{-0.4cm}
\caption{This figure shows the recovery of desired CoG position, linear and angular momentum during the balance board experiment.  X points in the robot right direction, Y in the forward direction and Z in the vertical direction. The grey shades are the moments when the board is moving, first up, then down. The total angle motion is from $-9.5^\circ$  to $9.5^\circ$ with respect to horizontal.}\label{fig:seesaw_momentum}
\vspace{-0.5cm}
\end{figure}

In an additional scenario the biped was standing on a balancing board. We ran the experiment with two configurations
for the robot: in one case the robot is standing such that the board motion happens in the sagittal plane and 
in the other case the motion happens in the lateral plane.
Fig.~\ref{fig:seesaw_momentum} shows results when the motion happens in the lateral plane.
In this case, the slope was varied in a range of $[-9.5^\circ; 9.5^\circ]$. Even for quite rapid changes in the slope, the feet remained flat on the ground. Compared to the push experiment the CoPs were moving in a wider range, but still remained in the interior of the foot soles with a margin. In this case, we notice a discrepancy in the predicted
contact forces and the real ones, making the case for a better dynamics model. Yet, the robot was still able to balance.

When we increased the pushes on the robot, eventually the momentum could not be damped out fast enough anymore and the robot reached a situation where the optimization could not find solutions anymore and the biped fell. In these cases the constraint that the feet have to be stationary was too restrictive. 
A higher level controller that takes into account stepping \cite{Pratt:2012ug,Urata:2012fb} becomes necessary to increase the stability margin.
As the balancing task required only two hierarchies, it took approximately \unit[0.4]{ms}  to solve the optimization problem at
each control cycle.%

\subsection{Tracking Experiments}
We also performed a tracking experiment. The goal was to track a desired CoM motion
while satisfying the physical constraints. The lower priority tasks consist of joint posture tracking
and contact forces regularization (i.e. we minimize tangential contact forces).
We wrote the CoM tracking task as a force task, using Eq.~\eqref{mom_rate_frcs} where the
desired CoM acceleration is computed using a PD controller.
The full hierarchy of tasks used in this experiment is shown in Table~\ref{tab:squatting_task}.

\begin{table}[]
\center
\begin{tabular}{c c l}
  \hline
  \textbf{Priority} & \textbf{Nr. of eq(uality) and} & \textbf{Constraint/Task} \\
	&  \textbf{ineq(uality) constraints} &\\
  \hline
  1 &$6$ eq& Newton Euler Eq.~\eqref{decomp_eq_motion_lower}\\
     &$2\times 14$ ineq& torque limits\\
 2 &$2 \times 6$ eq& kinematic contact constraint\\
     &$2\times 4$ ineq& CoPs reside in sup. polygons\\
     &$2\times 4$ ineq& GRFs reside in friction cones\\
     &$2 \times 14$ ineq& joint acceleration limits\\
 3 &$3$ eq& PD control on CoG\\
 4 &$14+6$ eq& PD control on posture\\
 5 &$2 \times 6$ eq& regularizer on GRFs\\
  \hline \hline
&\textbf{DoFs:} 14 &\textbf{max. time:}~ \unit[0.9]{ms} \bigstrut[t]\\
\end{tabular}
\vspace{-0.1cm}
\caption{\small Hierarchy of Tasks used in the Tracking Experiment}
\vspace{-0.4cm}
\label{tab:squatting_task}
\end{table}

\begin{figure}
\vspace{-0.0cm}
\centering
\subfigure[0.25 Hz high amplitude tracking task]{\includegraphics[width=0.35\textwidth]{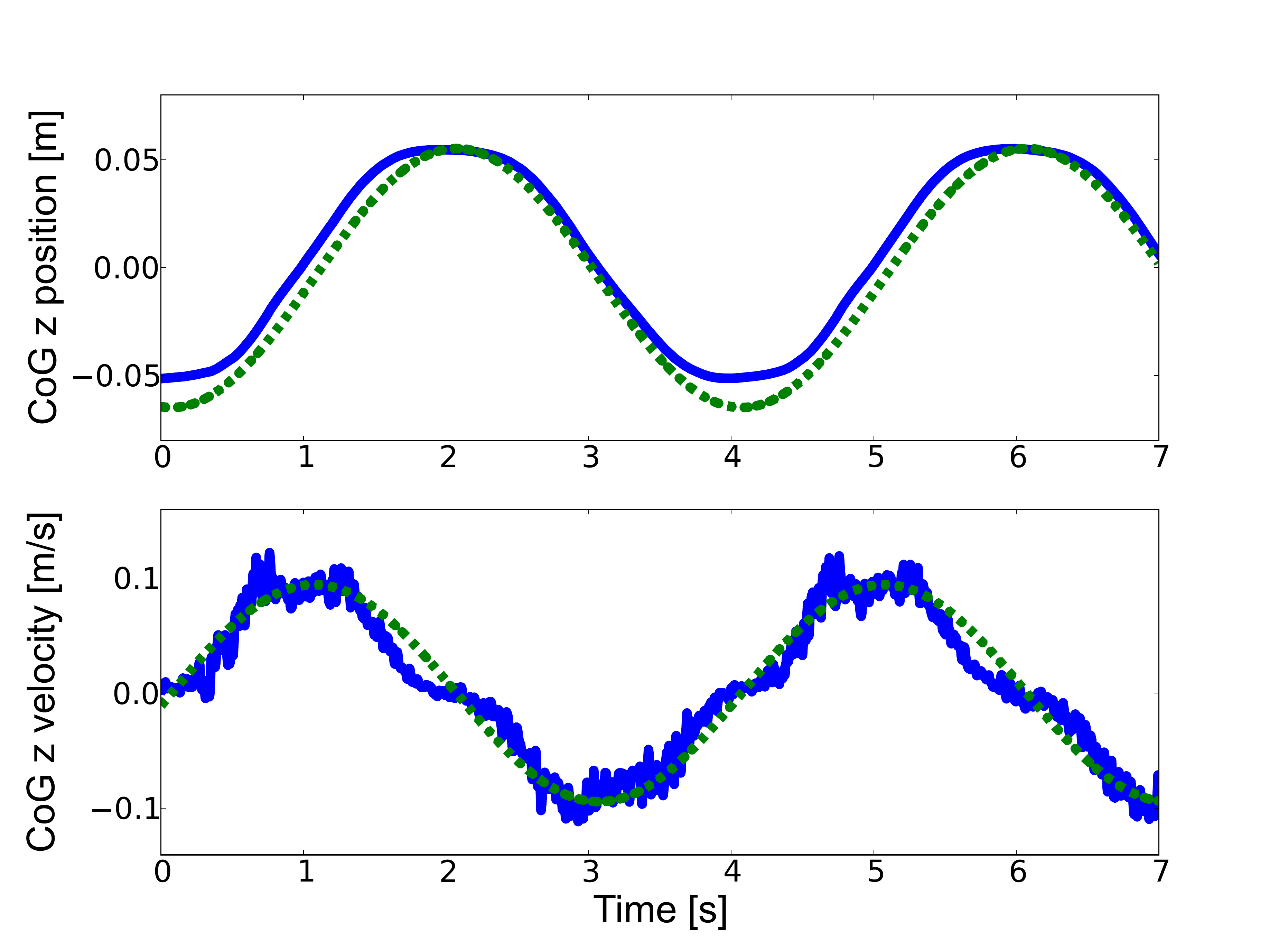}}
\subfigure[0.3 Hz low amplitude tracking task]{\includegraphics[width=0.35\textwidth]{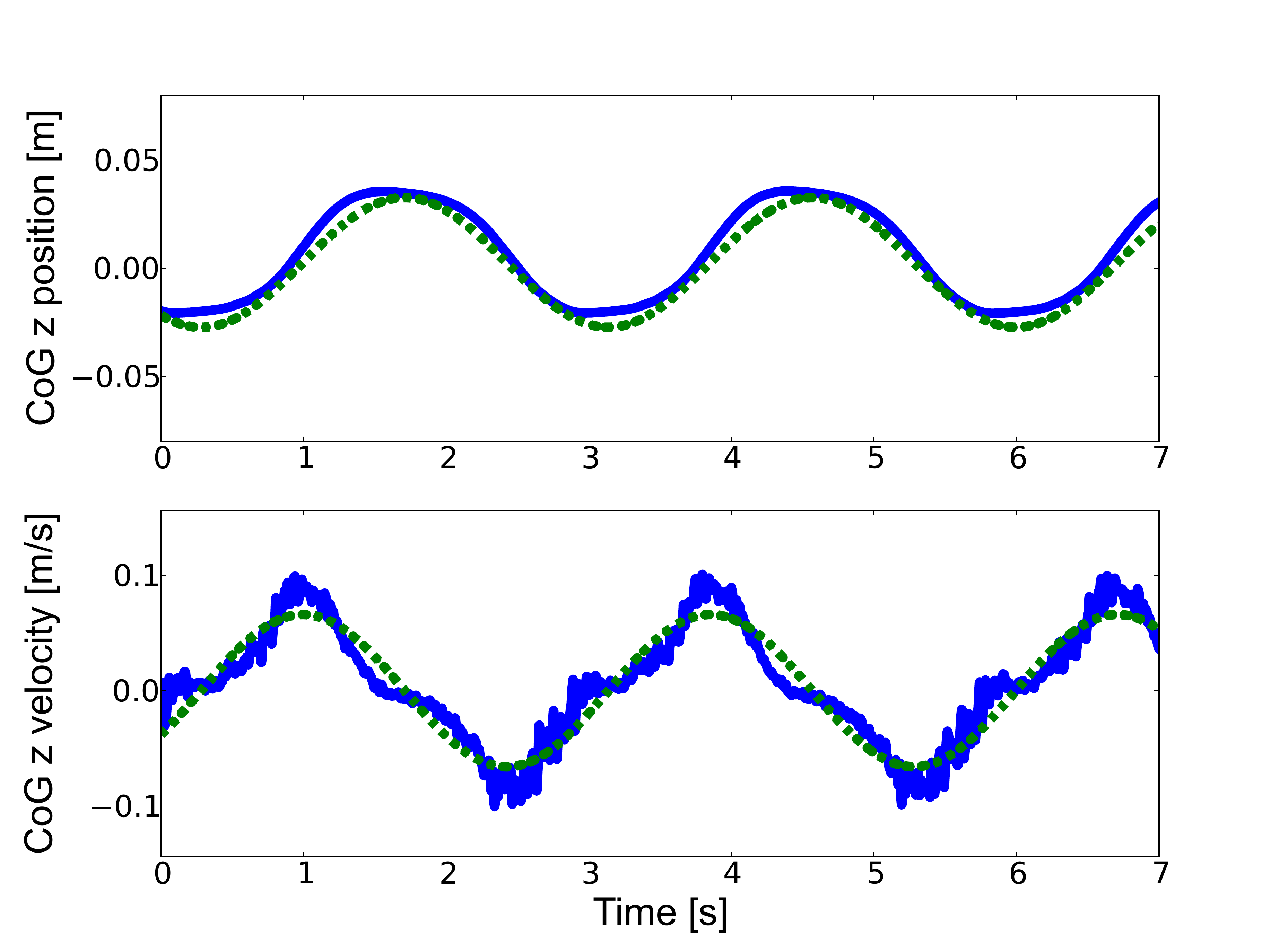}}
\vspace{-0.1cm}
\caption{Plot of the vertical CoM tracking performance for two tracking tasks. 
Both the positions (upper graphs) and velocities (lower graph) are shown in the plots. We note that both the desired positions and velocities (green dashed line) match very well the actual (blue plain line). The position RMSE is $0.0094\mathrm{m}$ for the upper plot and $0.0058$ for the lower plot.}\label{fig:com_tracking}
\vspace{-0.5cm}
\end{figure}

The CoM was tracking sine waves of various forms. The first one was a \unit[0.3]{Hz} sine wave
with \unit[0.02]{m} amplitude in the forward direction and \unit[0.03]{m} in the vertical direction. The second one
had a larger amplitude of \unit[0.06]{m} only in the vertical direction. 
Fig.~\ref{fig:com_tracking} shows the typical tracking performance during these tasks.
During the tracking experiments the robot was still able to handle a certain level of disturbances, as can
be seen in the attached video but not as much as when the angular momentum was also regulated.

In this experiment we use much more hierarchies than in the previous one. In this case, the controller
was able to compute the control commands in an average of \unit[0.9]{ms} (standard deviation of 0.045).
We see that in the current state, the hierarchical inverse dynamics is at the limit of the number of
hierarchies it can solve in the \unit[1]{kHz} control cycle.

\subsection{Balance in Single Support}
The experiments in the previous sections were done while the robot remained in double support. 
The goal of this experiment
is to show that the controller can handle more complicated tasks involving contact switching
and that the robot is able to balance on a single foot in face of disturbances.

The robot moves from a double support position to a single support position where the swing foot is lifted \unit[10]{cm} above the ground while balancing. This motion consists of 3 phases. First, the robot moves its CoM towards the center of the stance foot. Then an unloading phase occurs during which the contact force regularization enforces a zero contact force to guarantee  a continuous transition when the double support constraint is removed. In the final phase, a task controlling the motion of the swing foot is added to the hierarchy. 
Our contact switching strategy is simple but guarantees that continuous control commands were sent to the robot.
For more complicated tasks, such guarantees can always be met by using automatic task transitions 
such as in \cite{jarquin:humanoids:2013}.
The composition of hierarchies is summarized in Table~\ref{tab:one_foot}. Concerning computation time, the controller computes a solution in average well below 1ms but a maximum at 1.05ms is reached a few times during the unloading phase due to many constraints getting active at the same time.

\begin{figure}
\vspace{-0.4cm}
\centering
\includegraphics[width=0.4\textwidth]{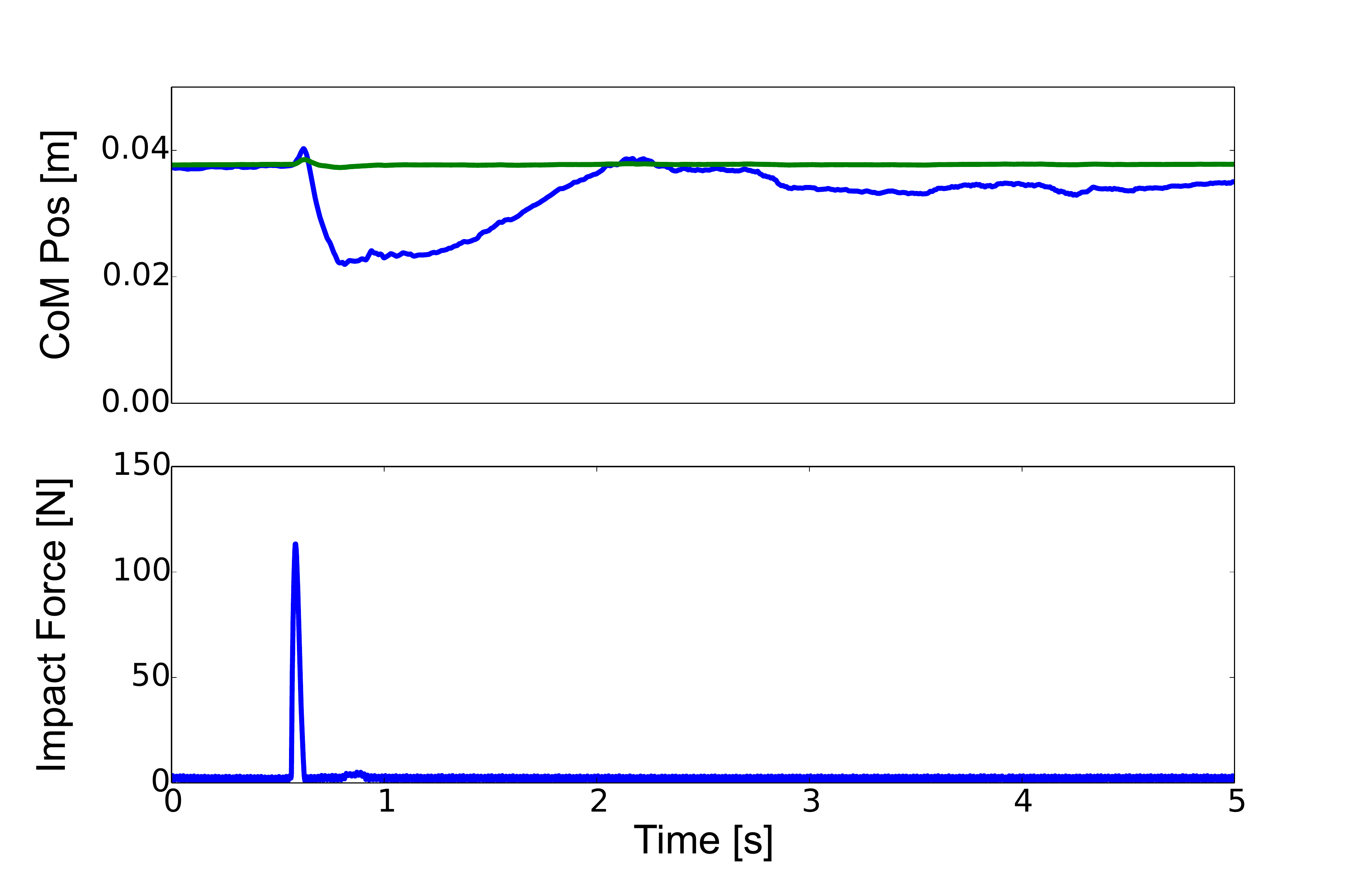}
\vspace{-0.2cm}
\caption{This plot shows the CoM tracking (upper graph) and the force exerted at the robot (lower graph) when it was balancing on one foot. The force was measured with a load cell that was attached to our push-stick. After the impulsive push the CoM (blue) deviated from the desired (green), but tracking was established again after a short duration.}
\label{fig:sing_bal_com}
\vspace{-0.0cm}
\end{figure}

Once in single support, we pushed the robot with a stick that has a force sensor attached to its tip allowing to measure the magnitude of the push. The robot was pushed impulsively with peak forces up to \unit[150]{N} and impulses between \unit[4.5]{Ns} and \unit[5.8]{Ns}, which compares nicely with other balance experiments \cite{Ott:2011uj}. Fig.~\ref{fig:sing_bal_com} shows the CoM tracking and exerted push force during a typical push. We can notice the fast recovery of the CoM position and the damped behavior which allows the CoM
to stay above support polygon. It is worth mentioning again that the foot size of the robot is rather small compared to other humanoids.

\begin{table}[]
\center
\begin{tabular}{c c l}
  \hline
  \textbf{Priority} & \textbf{Nr. of eq(uality) and} & \textbf{Constraint/Task} \\
	&  \textbf{ineq(uality) constraints} &\\
  \hline
  1 &$6$ eq& Newton Euler Eq.~\eqref{decomp_eq_motion_lower}\\
     &$2\times 14$ ineq& torque limits\\
 2 &$2\times 4$ ineq& CoPs reside in sup. polygons\\
     &$2\times 4$ ineq& GRFs reside in friction cones\\
     &$2 \times 14$ ineq& joint acceleration limits\\
 3 &$6$ eq. & Linear and angular momen-\\&&tum control\\
  &$12 / 6$ eq.& kinematic contact constraint \\
  &$0 / 6$ eq. & Cartesian foot motion (swing)\\
  &$14$ eq.& PD control on posture\\
 4 &$2 \times 6 / 1 \times 6$ eq.& regularizer on GRFs\\
  \hline \hline
&\textbf{DoFs:} 14 &\textbf{max. time:}~ \unit[1.05]{ms} \bigstrut[t]\\
\end{tabular}
\vspace{-0.2cm}
\caption{\small Hierarchy for Single Support Balancing experiment}
\label{tab:one_foot}
\vspace{-0.9cm}
\end{table}

\vspace{-0.1cm}
\section{DISCUSSION}
\vspace{-0.1cm}
In the following we discuss the results we presented and how they relate to other approaches.
\subsection{Relation to other balancing approaches}
The balance controller presented in the first experiment is a slight reformulation of the momentum-based controller
presented by Lee and Goswami \cite{Lee:2012hb,Lee:2010em}. Our formulation has the great
advantage to solve a single optimization problem instead of several ones and can therefore
guarantee that the control law will be consistent with all the constraints (joint limits, accelerations,
torque saturation, CoP limits and contact force limitations). Furthermore, we search over the full set of possible solutions and thus we are guaranteed to find the optimum, where~\cite{Lee:2012hb,Lee:2010em} optimize over sub-parts of the variables sequentially. It is also different from the work
of \cite{Kajita:2003gj} because we explicitly take into account contact forces in the optimization
and not purely kinematics.

The balance controller is very much related to the work of Stephens et al. \cite{Stephens:vu}.
In \cite{Stephens:vu}, the authors write the whole
optimization procedure using Eq. \eqref{equations_of_motion} with constraints similar to what we have.
However, the optimization problem being complicated, they actually solve a simpler problem were the contact
forces are first determined and then desired accelerations and torques are computed through a least-square solution.
From that point of view, torque saturation and limits on accelerations are not accounted for.
In our experiments, no tradeoff is necessary and we solve for all the constraints exactly. 
Further, the capability of strict task prioritization makes the design of more complicated tasks like balancing on one foot easier.

\subsection{Relations to other hierarchical inverse dynamics solvers}
In the hierarchical task solver that was used in this paper, we exploited the fact that the null space mapping of one priority could be computed in parallel with solving the QP of the same priority. On the other hand~\cite{Mansard:2012gy} suggest a more efficient way of dealing with inequalities by spending more computation time in constructing the problem. It is not clear which approach would be more efficient and a speed comparison would be very interesting. Their approach can also profit from the decomposition proposed in this paper, then it will not be required to compute an SVD of the full equations of motion, but only of the last six rows. In this paper our focus was the experimental evaluation of the discussed problem formulation, but as tasks become more complex and we will use the full robot, more efficient solvers will become necessary. 

\vspace{-0.cm}
\section{CONCLUSION}
\vspace{-0.1cm}
In this paper we have presented experimental results using a hierarchical inverse dynamics controller.
To the best of our knowledge, it is the first implementation of such an approach on a torque controlled
humanoid robot in a fast control loop. We have presented both balance and tracking experiments.
Our results suggest that the use of complete dynamic models and
hierarchies of tasks for feedback control is a feasible approach, despite the model inaccuracies and computational complexity. 
\vspace{-0.3cm}
\section*{ACKNOWLEDGMENT}
\vspace{-0.1cm}
We would like to thank Ambarish Goswami and  Seungkook Yun for hosting us at the Honda Research Institute
for one week and for their precious help in understanding the original momentum-based controller.
We would also like to thank Ambarish Goswami and Sung-Hee Lee for giving us an early access to their publication.
We are also grateful to Daniel Kappler for helping us with the videos, Nick Rotella for helping with the data acquisition from the IMU and Sean Mason for the help with the inertial parameters of the robot.

\vspace{0.6cm}
\bibliographystyle{IEEEtran}
{\footnotesize
\bibliography{icra2014}

\begin{thebibliography}{10}
\providecommand{\url}[1]{#1}
\csname url@rmstyle\endcsname
\providecommand{\newblock}{\relax}
\providecommand{\bibinfo}[2]{#2}
\providecommand\BIBentrySTDinterwordspacing{\spaceskip=0pt\relax}
\providecommand\BIBentryALTinterwordstretchfactor{4}
\providecommand\BIBentryALTinterwordspacing{\spaceskip=\fontdimen2\font plus
\BIBentryALTinterwordstretchfactor\fontdimen3\font minus
  \fontdimen4\font\relax}
\providecommand\BIBforeignlanguage[2]{{%
\expandafter\ifx\csname l@#1\endcsname\relax
\typeout{** WARNING: IEEEtran.bst: No hyphenation pattern has been}%
\typeout{** loaded for the language `#1'. Using the pattern for}%
\typeout{** the default language instead.}%
\else
\language=\csname l@#1\endcsname
\fi
#2}}

\bibitem{Hutter:2012uu}
M.~Hutter, M.~A. Hoepflinger, C.~Gehring, M.~Bloesch, C.~D. Remy, and
  R.~Siegwart, ``{Hybrid Operational Space Control for Compliant Legged
  Systems},'' in \emph{{R:SS}}, 2012.

\bibitem{Boaventura:2012tc}
T.~Boaventura, C.~Semini, J.~Buchli, M.~Frigerio, M.~Focchi, and D.~Caldwell,
  ``{Dynamic Torque Control of a Hydraulic Quadruped Robot},'' in
  \emph{{ICRA}}, 2012.

\bibitem{Cheng:2008tw}
G.~Cheng, H.~Sang-Ho, A.~Ude, J.~Morimoto, J.~G. Hale, J.~Hart, J.~Nakanishi,
  D.~Bentivegna, J.~Hodgins, C.~Atkeson, M.~Mistry, S.~Schaal, and M.~Kawato,
  ``{CB: Exploring Neuroscience with a Humanoid Research Platform},'' in
  \emph{{ICRA}}, 2008.

\bibitem{Ott:2011uj}
C.~Ott, M.~A. Roa, and G.~Hirzinger, ``{Posture and balance control for biped
  robots based on contact force optimization},'' in \emph{{Humanoids}}, 2011.

\bibitem{Kalakrishnan:2011dy}
M.~Kalakrishnan, J.~Buchli, P.~Pastor, M.~Mistry, and S.~Schaal, ``{Learning,
  Planning, and Control for Quadruped Locomotion over Challenging Terrain},''
  \emph{{IJRR}}, vol.~30, no.~2, pp. 236--258, 2011.

\bibitem{Saab:2011gg}
L.~Saab, O.~Ramos, N.~Mansard, P.~Soueres, and J.-Y. Fourquet, ``{Generic
  Dynamic Motion Generation with Multiple Unilateral Constraints},'' in
  \emph{{IROS}}, 2011.

\bibitem{Salini:2011bw}
J.~Salini, V.~Padois, and P.~Bidaud, ``{Synthesis of Complex Humanoid
  Whole-Body Behavior: A Focus on Sequencing and Tasks Transitions},'' in
  \emph{{ICRA}}, 2011.

\bibitem{Righetti:2012uc}
L.~Righetti and S.~Schaal, ``{Quadratic programming for inverse dynamics with
  optimal distribution of contact forces},'' in \emph{{Humanoids}}, 2012.

\bibitem{Righetti:2013tt}
L.~Righetti, J.~Buchli, M.~Mistry, M.~Kalakrishnan, and S.~Schaal, ``{Optimal
  distribution of contact forces with inverse-dynamics control},''
  \emph{{IJRR}}, vol.~32, no.~3, pp. 280--298, 2013.

\bibitem{Hyon:2007jya}
S.-H. Hyon, J.~G. Hale, and G.~Cheng, ``{Full-Body Compliant Human--Humanoid
  Interaction: Balancing in the Presence of Unknown External Forces},''
  \emph{Transactions on Robotics}, vol.~23, no.~5, pp. 884--898, 2007.

\bibitem{Lee:2012hb}
S.-H. Lee and A.~Goswami, ``{A momentum-based balance controller for humanoid
  robots on non-level and non-stationary ground},'' \emph{Autonomous Robots},
  vol.~33, pp. 399--414, 2012.

\bibitem{Stephens:vu}
B.~J. Stephens and C.~G. Atkeson, ``{Dynamic Balance Force Control for
  compliant humanoid robots},'' in \emph{{IROS}}, 2010.

\bibitem{deLasa:2010hf}
M.~de~Lasa, I.~Mordatch, and A.~Hertzmann, ``Feature-{B}ased {L}ocomotion
  {C}ontrollers,'' \emph{ACM Transactions on Graphics}, vol.~29, no.~3, 2010.

\bibitem{Mansard:2012gy}
N.~Mansard, ``{A dedicated solver for fast operational-space inverse
  dynamics},'' in \emph{{ICRA}}, 2012.

\bibitem{Kajita:2003gj}
S.~Kajita, F.~Kanehiro, K.~Kaneko, K.~Fujiwara, K.~Harada, K.~Yokoi, and
  H.~Hirukawa, ``{Resolved momentum control: humanoid motion planning based on
  the linear and angular momentum},'' in \emph{{IROS}}, 2003.

\bibitem{Wieber:2006up}
P.~Wieber, ``{Holonomy and nonholonomy in the dynamics of articulated
  motion},'' \emph{Fast motions in biomechanics and robotics}, pp. 411--425,
  2006.

\bibitem{herzog:2013a}
A.~Herzog, L.~Righetti, F.~Grimminger, P.~Pastor, and S.~Schaal,
  ``Momentum-based balance control for torque-controlled humanoids,'' vol.
  http://arxiv.org/abs/1305.2042v1, 2013.

\bibitem{Boaventura:2012va}
T.~Boaventura, M.~Focchi, M.~Frigerio, J.~Buchli, C.~Semini, G.~A.
  Medrano-Cerda, and D.~Caldwell, ``{On the role of load motion compensation in
  high-performance force control},'' in \emph{{IROS}}, 2012.

\bibitem{Mistry:2009dh}
M.~Mistry, S.~Schaal, and K.~Yamane, ``{Inertial Parameter Estimation of
  Floating Base Humanoid Systems using Partial Force Sensing},'' in
  \emph{{Humanoids}}, 2009.

\bibitem{Bloesch:2012wu}
M.~Bloesch, M.~Hutter, M.~H. Hoepflinger, C.~D. Remy, C.~Gehring, and
  R.~Siegwart, ``{State estimation for legged robots-consistent fusion of leg
  kinematics and IMU},'' in \emph{{R:SS}}, 2012.

\bibitem{Orin:2008ge}
D.~E. Orin and A.~Goswami, ``{Centroidal Momentum Matrix of a humanoid robot:
  Structure and Properties},'' in \emph{{IROS}}, 2008.

\bibitem{Pratt:2012ug}
J.~Pratt, T.~Koolen, T.~De~Boer, J.~Rebula, S.~Cotton, J.~Carff, M.~Johnson,
  and P.~Neuhaus, ``{Capturability-based analysis and control of legged
  locomotion, Part 2: Application to M2V2, a lower-body humanoid},''
  \emph{{IJRR}}, vol.~31, no.~10, 2012.

\bibitem{Urata:2012fb}
J.~Urata, K.~Nshiwaki, Y.~Nakanishi, K.~Okada, S.~Kagami, and M.~Inaba,
  ``{Online Walking Pattern Generation for Push Recovery and Minimum Delay to
  Commanded Change of Direction and Speed},'' in \emph{{IROS}}, 2012.

\bibitem{jarquin:humanoids:2013}
G.~Jarquin, A.~Escande, G.~Arechavaleta, T.~Moulard, E.~Yoshida, and
  V.~Parra-Vega, ``Real-time smooth task transitions for hierarchical inverse
  kinematics,'' in \emph{{Humanoids}}, 2013.

\bibitem{Lee:2010em}
S.-H. Lee and A.~Goswami, ``{Ground reaction force control at each foot: A
  momentum-based humanoid balance controller for non-level and non-stationary
  ground},'' in \emph{{IROS}}, 2010, pp. 3157--3162.

\end{thebibliography}
}

\end{document}